\pdfoutput=1
\documentclass[11pt]{article}

\usepackage[]{ACL2023}
\usepackage{times}
\usepackage{latexsym}
\usepackage[T1]{fontenc}
\usepackage[utf8]{inputenc}
\usepackage{microtype}
\usepackage{xcolor}
\usepackage{algorithmic}
\usepackage{inconsolata}
\usepackage{amsmath,amsfonts,bm}

\def\eqref#1{equation~\ref{#1}}
\def\1{\bm{1}}

\DeclareMathAlphabet{\mathsfit}{\encodingdefault}{\sfdefault}{m}{sl}
\SetMathAlphabet{\mathsfit}{bold}{\encodingdefault}{\sfdefault}{bx}{n}

\usepackage{threeparttable}
\usepackage{multirow}
\usepackage{colortbl}
\definecolor{mygray}{gray}{0.85}

\usepackage{xspace}
\newcommand{\model}{CFPRM\xspace}

\newcommand{\vpara}[1]{\vspace{0.04in}\noindent\textbf{#1}\xspace}

\usepackage{subfigure}
\usepackage{multirow}
\usepackage[normalem]{ulem}
\usepackage[ruled,linesnumbered]{algorithm2e}

\usepackage{xspace}
\usepackage{threeparttable}
\usepackage{enumitem}
\usepackage{makecell}
\usepackage{bbding}
\usepackage{caption}
\usepackage{graphics}
\usepackage{graphicx}
\usepackage{CJKutf8}
\usepackage{subcaption} 
\usepackage{subfigure}

\usepackage{hyperref}

\usepackage[utf8]{inputenc} % allow utf-8 input
\usepackage[T1]{fontenc}    % use 8-bit T1 fonts
\usepackage{url}            % simple URL typesetting
\usepackage{booktabs}       % professional-quality tables
\usepackage{amsfonts}       % blackboard math symbols
\usepackage{nicefrac}       % compact symbols for 1/2, etc.
\usepackage{microtype}      % microtypography
\usepackage{lipsum}
\usepackage{csquotes}
\usepackage{fancyhdr}       % header
\usepackage{longtable}
\usepackage{graphicx}       % graphics
\usepackage{multirow}
\usepackage{booktabs}

\title{Coarse-to-Fine Process Reward Modeling for Mathematical Reasoning}
\author{\bf
Yulan Hu$^{1}$,
Sheng Ouyang$^{1}$,
Jinman Zhao$^2$,
Yong Liu$^1$
\\
$^1$ Renmin University of China, Gaoling School of Artificial Intelligence \\
$^2$ 	University of Toronto\\
\texttt{huyulan,ouyangsheng@ruc.edu.cn}\\
}

\begin{document}
\maketitle
\begin{abstract}
The Process Reward Model (PRM) plays a crucial role in mathematical reasoning tasks, requiring high-quality supervised process data. However, we observe that reasoning steps generated by Large Language Models (LLMs) often fail to exhibit strictly incremental information, leading to redundancy that can hinder effective reasoning. To address this issue, we propose \model, a simple yet effective coarse-to-fine strategy. Instead of focusing on the detection of redundant steps, our approach first establishes a coarse-grained window to merge adjacent reasoning steps into unified, holistic steps. The window size is then progressively reduced to extract fine-grained reasoning steps, enabling data collection at multiple granularities for training. By leveraging this hierarchical refinement process, \model mitigates redundancy while preserving essential fine-grained knowledge. Extensive experiments on two reasoning datasets across three loss criteria validate the \model's effectiveness and versatility. Our code will be released in the future.
\end{abstract}

\section{Introduction}
Large language models (LLMs) have demonstrated promising capabilities across a wide range of domains~\cite{llm_applications0,gpt4,llama3.1,qwen2.5}, including complex mathematical reasoning tasks~\cite{verify_step_by_step,huang2023large}. An accurate process reward model (PRM) is vital for reasoning tasks, as it provides intermediate supervision signals for each individual step~\cite{uesato2022solving}.

\begin{figure}[htb]
    \centering
    \includegraphics[width=1.0\linewidth]{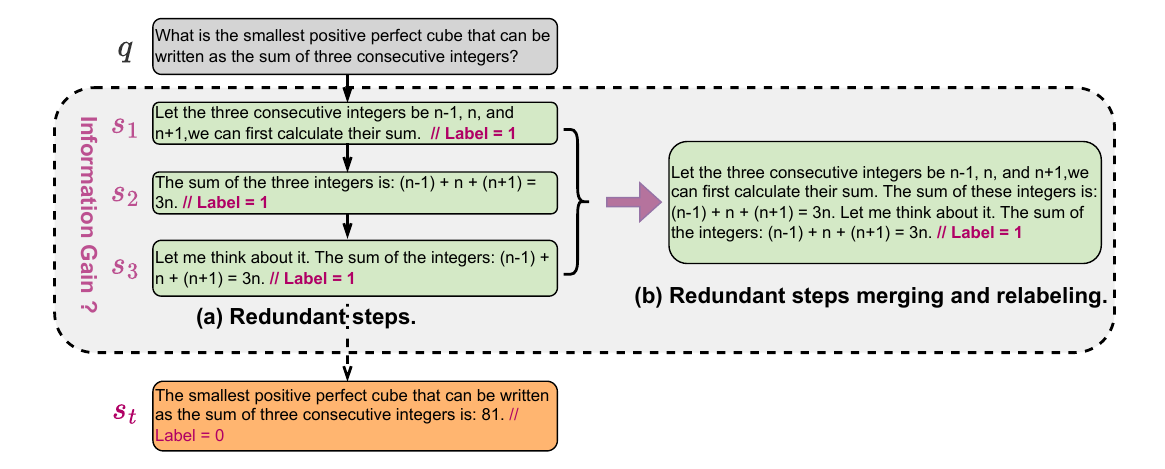}
    \caption{Redundant steps merging.}
    \label{fig:annotation_example}
\end{figure}

Training PRM requires the collection of step-wise annotated corpora~\cite{verify_step_by_step, uesato2022solving}. For instance, Lightman et al.~\cite{verify_step_by_step} propose manually annotating the intermediate MATH data, where each step is assigned a ternary label. However, such human-intensive labeling is costly, hindering broader practical applications. An alternative approach involves constructing automatic labeling methods, either by defining the probability of each intermediate step as the potential to deduce the final correct answer~\cite{shepherd}, or by using a tree-based structure to iteratively refine the logits of each intermediate trajectory~\cite{restmcts}. Despite the preliminary success of these methods, they primarily focus on accurately assigning labels to each step, while overlooking the potential redundancy of steps that may offer no incremental information gain~\cite{pqm}. Given that mathematical reasoning is a progressive process, where each current step depends on previous ones~\cite{pqm}, later steps should ideally provide more informative contributions toward approximating the final answer. To illustrate this, we present a data collection example from the MATH dataset~\cite{math_dataset} via ShepHerd~\cite{shepherd} in Figure~\ref{fig:annotation_example}. However, we observe that steps $s_{1}$, $s_{2}$, and $s_{3}$ are logically correct, but the repetitive reasoning procedures fail to yield any new information, which contradicts the learning objective.

To tackle the limitation, we propose \model, a coarse-to-fine strategy for process data collection and training, which is simple yet effective. We do not explicitly detect redundant steps; as the name suggests, we collect process training data in a coarse-to-fine manner and proceed with the learning process in the same way. Specifically, we define a step window size $C$ to represent the initial step granularity, i.e., every $C$ steps are collected and merged into a holistic step, with the corresponding label of the merged step determined by the label of the last individual step. Subsequently, $C$ is gradually reduced until it reaches 1, and training data are collected in the same way following the above procedure. This strategy gathers training data of diverse granularity, directly integrating consecutive steps to form coarse steps without designing methods to detect redundant steps. Meanwhile, the initial individual steps are preserved to offer necessary fine-grained signals. We validate the proposed strategy on two cutting-edge LLMs across three learning criteria, yielding consistently enhanced performance, demonstrating the effectiveness and versatility of \model.

\section{Methodology}
\subsection{Preliminaries}~\label{sec:preliminaries}
We denote an LLM policy as $\pi$, and $r_\theta$ as the PRM fine-tuned upon $\pi$, parameterized by $\theta$. For reasoning tasks, $\pi$ generates responses step by step given an input query $x$ in an autoregressive manner: $s_t \sim \pi_\theta (\cdot \mid x, s_{1:t-1}), t \leq T$, $T$ is the total reasoning steps. The PRM policy $r_\theta$ then outputs a reward given the partial solutions and the input query as: \( r_{s_t} = r_\theta(s_{1:t}, x) \). We regard \( y_{s_t} \) as the label for step \( t \). In addition, the existing PRM training objectives can be summarized into three types, including mean square error (MSE)~\cite{restmcts}, binary cross-entropy (BCE)~\cite{shepherd}, and Q-value rankings (Q-ranking)~\cite{pqm}, as shown in Table~\ref{equa:loss}. It is worth noting that \model can be applied to arbitrary loss criteria and achieves consistent improvements, as demonstrated in Section~\ref{sec:experiments}.

\begin{table}[h]
\caption{The typical loss objectives for PRM training.}
\resizebox{\linewidth}{!}{
\begin{tabular}{@{}c|l@{}}
\toprule
Losses          & Formulation                                                                                                                                                                                                                                                                                                                                                                                                                                    \\ \midrule
BCE             & \( \sum_{t=1}^T y_{s_t} \log r_{s_t} + \left(1 - y_{s_t}\right) \log \left(1 - r_{s_t}\right) \)                                                                                                                                                                                                        \\ \midrule
MSE             & \( \sum_{t=1}^T \left(r_\theta(s_{1:t}, x) - y_{s_t}\right)^2 \)                                                                                                                                                                                                                                                                        \\ \midrule
Q-Ranking & \( -\frac{1}{|T|} \sum_{t=0}^{|T|} \log \frac{\exp \left(r_{c_t}\right)}{\sum_{q=0}^t \exp r_{c_q} + \sum_{w \in W} \exp \left(Q_w + \zeta\right)} \)  \\ \bottomrule
\end{tabular}
}
\begin{tablenotes}
    \footnotesize
    \item[] \scriptsize $W$ indicates negative steps, $Q$ is the ranking value, and $\zeta$ is the margin hyperparameter.
    \end{tablenotes}
\label{equa:loss}
\end{table}

\subsection{Coarse-to-fine Process Data Collection}~\label{sec:cfprm_main}
\begin{figure}[h]
    \centering
    \includegraphics[width=1.\linewidth]{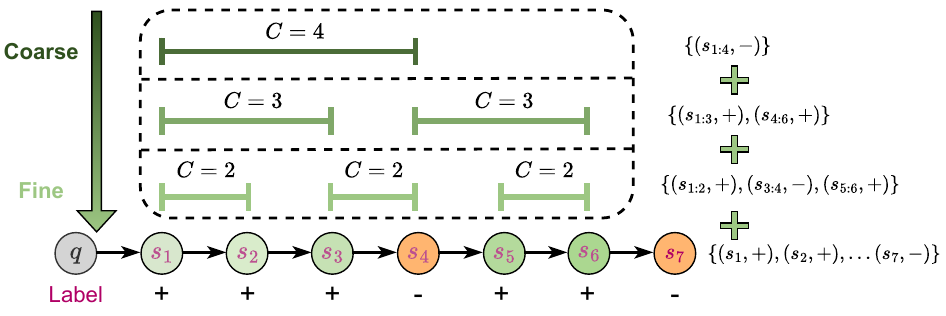}
    \caption{Coarse-to-fine process data collection.}
    \label{fig:coarse_to_fine}
\end{figure}

\model can be applied to reasoning data collected by any kind of structure, such as Chain of Thought~\cite{shepherd} or Monte Carlo Tree Search~\cite{restmcts} methods. We extend the solution process in Figure~\ref{fig:annotation_example} as an example and present the coarse-to-fine process data collection in Figure~\ref{fig:coarse_to_fine}. The process is intuitive, involving first merging the consecutive steps and then relabeling each merged step.

\begin{table*}[h]
\caption{Main results measured by BoN accuracy. $C$ is set to 2.}
\resizebox{\linewidth}{!}{
\begin{tabular}{@{}c|c|ccccc|ccccc@{}}
\toprule
\multirow{2}{*}{Version} & \multirow{2}{*}{Models} & \multicolumn{4}{c}{GSM-Plus} &      & \multicolumn{4}{c}{MATH500} &      \\ \cmidrule(l){3-12} 
                                                                                &                         & @8    & @16   & @32   & @64  & \cellcolor[HTML]{cee4ff}Avg. & @8    & @16   & @32  & @64  & \cellcolor[HTML]{cee4ff}Avg. \\ \midrule
\multirow{7}{*}{Instruct}                                                       & ORM                     & 67.0  & 66.0  & 68.8  & 66.8 & \cellcolor[HTML]{cee4ff}67.2 & 71.4  & 69.2  & 70.2 & 68.1 & \cellcolor[HTML]{cee4ff}69.7 \\ \cmidrule(l){2-12} 
                                                                                & SHerd                     & 67.8  & 67.2  & 68.4  & 67.0 & \cellcolor[HTML]{cee4ff}67.6 & 74.4  & 75.5  & 75.8 & 76.0 & \cellcolor[HTML]{cee4ff}75.4 \\
                                                                                & $\text{SHerd}_{+\text{\model}}$                    & 68.2  & 67.4  & 69.0  & 70.0 & \cellcolor[HTML]{cee4ff}$\text{\textbf{68.7}}_{\textcolor{blue}{\uparrow 1.1}}$ & 75.6  & 76.2  & 76.6 & 77.0 & \cellcolor[HTML]{cee4ff}$\text{\textbf{76.4}}_{\textcolor{blue}{\uparrow 1.0}}$  \\ \cmidrule(l){2-12} 
                                                                                & RMCTS*                     & 69.4  & 69.2  & 66.8  & 68.0 & \cellcolor[HTML]{cee4ff}68.4 & 71.0  & 71.2  & 71.9 & 72.6 & \cellcolor[HTML]{cee4ff}71.7 \\
                                                                                & $\text{RMCTS*}_{+\text{\model}}$                    & 68.2  & 70.4  & 70.2  & 70.0 & \cellcolor[HTML]{cee4ff}$\text{\textbf{69.7}}_{\textcolor{blue}{\uparrow 1.3}}$ & 75.2  & 75.6  & 74.9 & 74.6 & \cellcolor[HTML]{cee4ff}$\text{\textbf{75.1}}_{\textcolor{blue}{\uparrow 3.4}}$ \\ \cmidrule(l){2-12} 
                                                                                & PQM & 67.6  & 68.8  & 66.4  & 67.0 & \cellcolor[HTML]{cee4ff}67.5 & 74.6  & 75.4  & 75.8 & 75.3 & \cellcolor[HTML]{cee4ff}75.3 \\
                                                                                & $\text{PQM}_{+\text{\model}}$                    & 68.0  & 68.0  & 69.4  & 71.0 & \cellcolor[HTML]{cee4ff}$\text{\textbf{69.1}}_{\textcolor{blue}{\uparrow 1.6}}$ & 75.4  & 76.2  & 76.7 & 77.1 & \cellcolor[HTML]{cee4ff}$\text{\textbf{76.4}}_{\textcolor{blue}{\uparrow 1.1}}$  \\ \midrule \midrule
\multirow{7}{*}{MATH}                                                           & ORM                     & 63.0  & 61.8  & 62.8  & 62.8 & \cellcolor[HTML]{cee4ff}62.6 & 78.0  & 77.7  & 77.6 & 77.6 & \cellcolor[HTML]{cee4ff}77.7 \\ \cmidrule(l){2-12} 
                                                                                & SHerd                     & 69.2  & 69.2  & 69.2  & 70.0 & \cellcolor[HTML]{cee4ff}69.4 & 82.1  & 81.8  & 81.9 & 82.0 & \cellcolor[HTML]{cee4ff}82.0 \\
                                                                                & $\text{SHerd}_{+\text{\model}}$                    & 69.2  & 70.8 & 71.2 & 73.0 & \cellcolor[HTML]{cee4ff}$\text{\textbf{71.2}}_{\textcolor{blue}{\uparrow 1.8}}$ & 82.4  & 82.3  & 82.8 & 82.8 & \cellcolor[HTML]{cee4ff}$\text{\textbf{82.6}}_{\textcolor{blue}{\uparrow 0.6}}$  \\ \cmidrule(l){2-12} 
                                                                                & RMCTS*                     & 68.6  & 69.2  & 68.6  & 70.0 & \cellcolor[HTML]{cee4ff}69.1 & 81.6  & 81.7  & 82.0 & 81.8 & \cellcolor[HTML]{cee4ff}81.8 \\
                                                                                & $\text{RMCTS*}_{+\text{\model}}$  & 69.2  & 70.0  & 70.4  & 72.0 & \cellcolor[HTML]{cee4ff}$\text{\textbf{70.4}}_{\textcolor{blue}{\uparrow 1.3}}$ & 82.2  & 82.7  & 82.8 & 82.8 & \cellcolor[HTML]{cee4ff}$\text{\textbf{82.6}}_{\textcolor{blue}{\uparrow 0.8}}$ \\ \cmidrule(l){2-12} 
                                                                                & PQM  & 70.2  & 69.8  & 72.2  & 73.0 & \cellcolor[HTML]{cee4ff}71.3 & 84.0  & 84.1  & 84.2 & 84.2 & \cellcolor[HTML]{cee4ff}84.1 \\
                                                                                & $\text{PQM}_{+\text{\model}}$  & 69.4  & 71.4  & 72.2  & 74.0 & \cellcolor[HTML]{cee4ff}$\text{\textbf{71.8}}_{\textcolor{blue}{\uparrow 0.5}}$ & 84.5  & 84.9  & 85.2 & 85.2 & \cellcolor[HTML]{cee4ff}$\text{\textbf{85.0}}_{\textcolor{blue}{\uparrow 0.9}}$ \\ \bottomrule
\end{tabular}}
\label{tab:main_results}
\end{table*}

\vpara{Steps merging.} Consider the problem in Figure~\ref{fig:annotation_example} containing a trajectory of 7 reasoning steps, where steps \( s_1 \), \( s_2 \), \( s_3 \), \( s_5 \), and \( s_6 \) are correct steps, and \( s_4 \) and \( s_7 \) are wrong steps. It is worth noting that step \( s_4 \) fails to make incremental reasoning from \( s_3 \), but the LLM policy manages to adjust the wrong step~\cite{pav}. The final step fails to reach the correct answer.

As a coarse-to-fine method, \model gradually consolidates multiple reasoning steps based on a predefined sliding window size, denoted as \( C \). Here, \( C \) represents the size of the merging window, with \( C_{\text{max}} \) as the maximum window size, while the minimum size is usually set to 1. Ideally, \( C \) can initially be set to the total number of steps in the entire reasoning trajectory, and the merged trajectory is labeled with the label of the last step, acting as ORM. In practice, the continuous partial steps are merged instead of the entire trajectory to avoid excessive concentration of knowledge. In Figure~\ref{fig:coarse_to_fine}, \( C \) is initially set to 4, yielding the merged partial trajectory \( s_1, s_2, s_3, s_4 \), treated as a single step \( s_{1:4} \). Then, the window slides to the next starting point (\( s_5 \)) to collect a new partial trajectory. Subsequently, \( C \) is sequentially decreased to 3 and 2, yielding additional partial trajectories of different sizes. After finishing collecting the consecutive steps using the sliding window, we combine the merged coarse training samples with the original fine-grained data.

\vpara{Merged steps labeling.} In addition, each collected partial trajectory may contain positive or negative steps, which makes it difficult to determine its label. Drawing inspiration from ShepHerd~\cite{shepherd}, which considers a trajectory's label to depend on its potential to deduce the answer, we label each merged step by the label of the last step in the window. For example, the label of \( s_{1:4} \) is the same as the label of step \( s_4 \), indicating a negative sample, and we add \( (s_{1:4}, +) \) into the set \( \mathcal{D}_{C=4} \). Similarly, the trajectory \( s_{1:2} \) is treated as a positive sample, sharing the same label as step \( s_2 \). Following this, each merged step is relabeled and used as supervisory training samples, with the trajectory added to the set \( \mathcal{D}_{C=2} \). In this way, we obtain a renewed training corpus containing samples of diverse granularity.

\vpara{Training and inference.} We proceed with training after obtaining the training samples of different granularity. We choose to recombine the training trajectories according to their granularity. Specifically, we traverse the corpus sequentially from \( C_{\text{max}} \) to 1. Through this process, the process supervision knowledge is gradually distilled in a coarse-to-fine manner. The training process is identical for different loss criteria. During inference, we use the trained PRM to predict scores for each single step. We summarize the overall process in Appendix~\ref{algor}.

\section{Experiment}~\label{sec:experiments}
\subsection{Experimental Setup}
\vpara{Datasets and models.} We adopt two widely used mathematical reasoning test sets, GSM-Plus~\cite{gsmplus} and MATH500~\cite{math_dataset}, for evaluation. GSM-Plus is built upon GSM8K~\cite{gsm8k} with various mathematical perturbations. The original GSM8K is a benchmark of grade-school level problems. The MATH dataset consists of high school math competition problems, which are more challenging. For each candidate, the PRM evaluates the score of each step. Instead of collecting raw process data from scratch, we utilize the off-the-shelf PRM800K~\cite{verify_step_by_step} dataset to train the PRM. We employ two cutting-edge LLMs, Qwen2.5-7B-Instruct and Qwen2.5-7B-MATH~\cite{qwen2.5}, as the backbone models. Following previous studies~\cite{shepherd,pqm}, we evaluate the PRM using the best-of-$n$ (BoN) sampling strategy, with $n$ set to 8, 16, 32, and 64, respectively. We also use the same backbone model to generate 64 candidates for each given question, ensuring consistency in the backbone model and the BoN sampling policy.

\vpara{Baselines and details.} As a method to refine the data collection mechanism, \model can be seamlessly applied to any existing methods. Specifically, we select the most recent methods covering the three loss criteria in Table~\ref{equa:loss} for comparison, including ShepHerd~\cite{shepherd} using the BCE objective, ReSTMCTS*~\cite{restmcts} using the MSE objective, and PQM~\cite{pqm} built upon the Q-ranking objective. We also include ORM for comparison. The best performance is marked in bold. We set the max length to 2048, the learning rate to 2e-6, and the batch size to 32. All experiments are conducted on an H800-80G GPU, with each experiment repeated five times to report the mean results. Accuracy is reported as the evaluation metric.

\begin{figure}[h]    
  \centering
  \subfigure[BCE on GSM-Plus.]{
    \label{local_sample_construction}
    \includegraphics[width=0.23\textwidth]{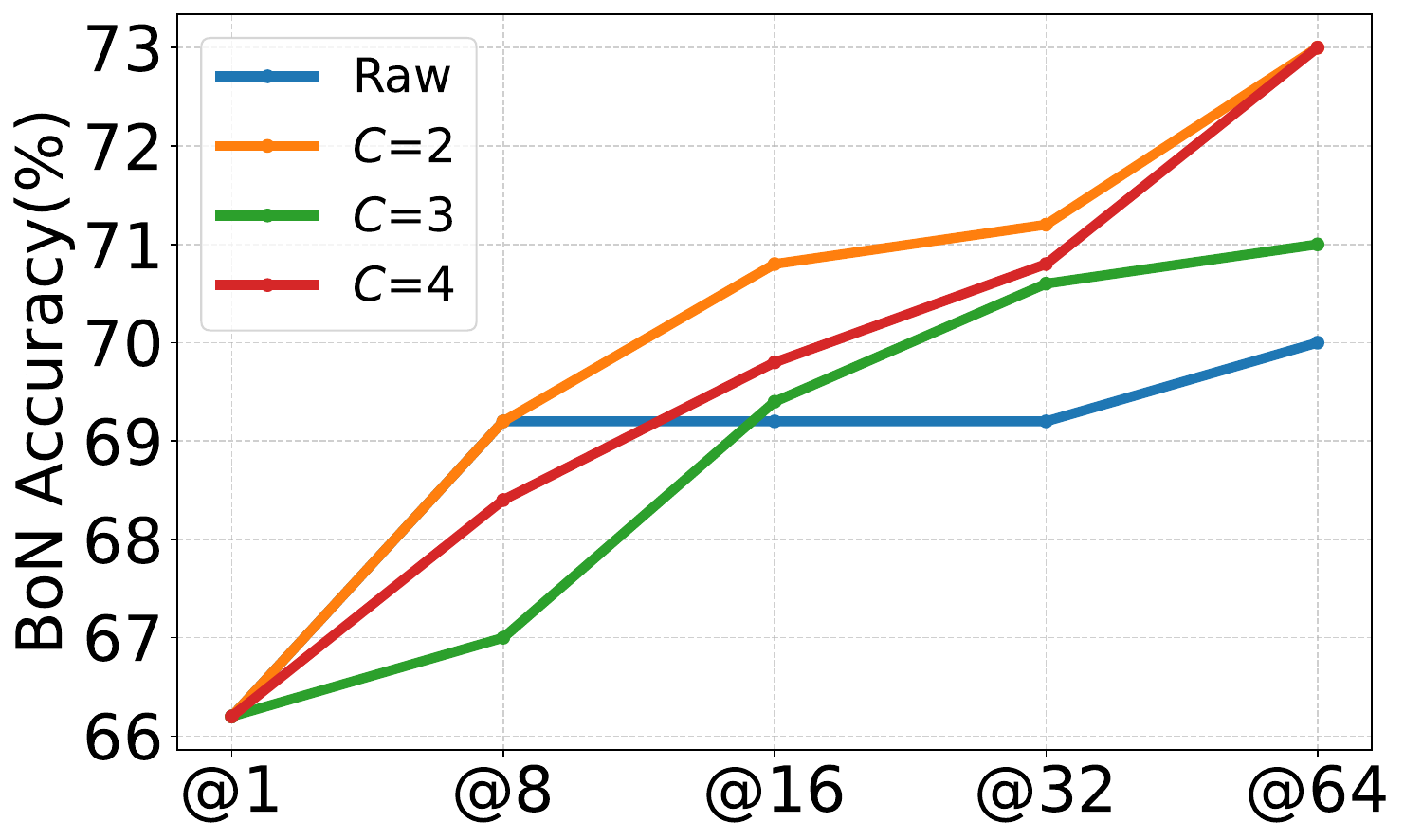}}
  \subfigure[BCE on MATH500.]{
    \label{subgraph_instance}
    \includegraphics[width=0.23\textwidth]{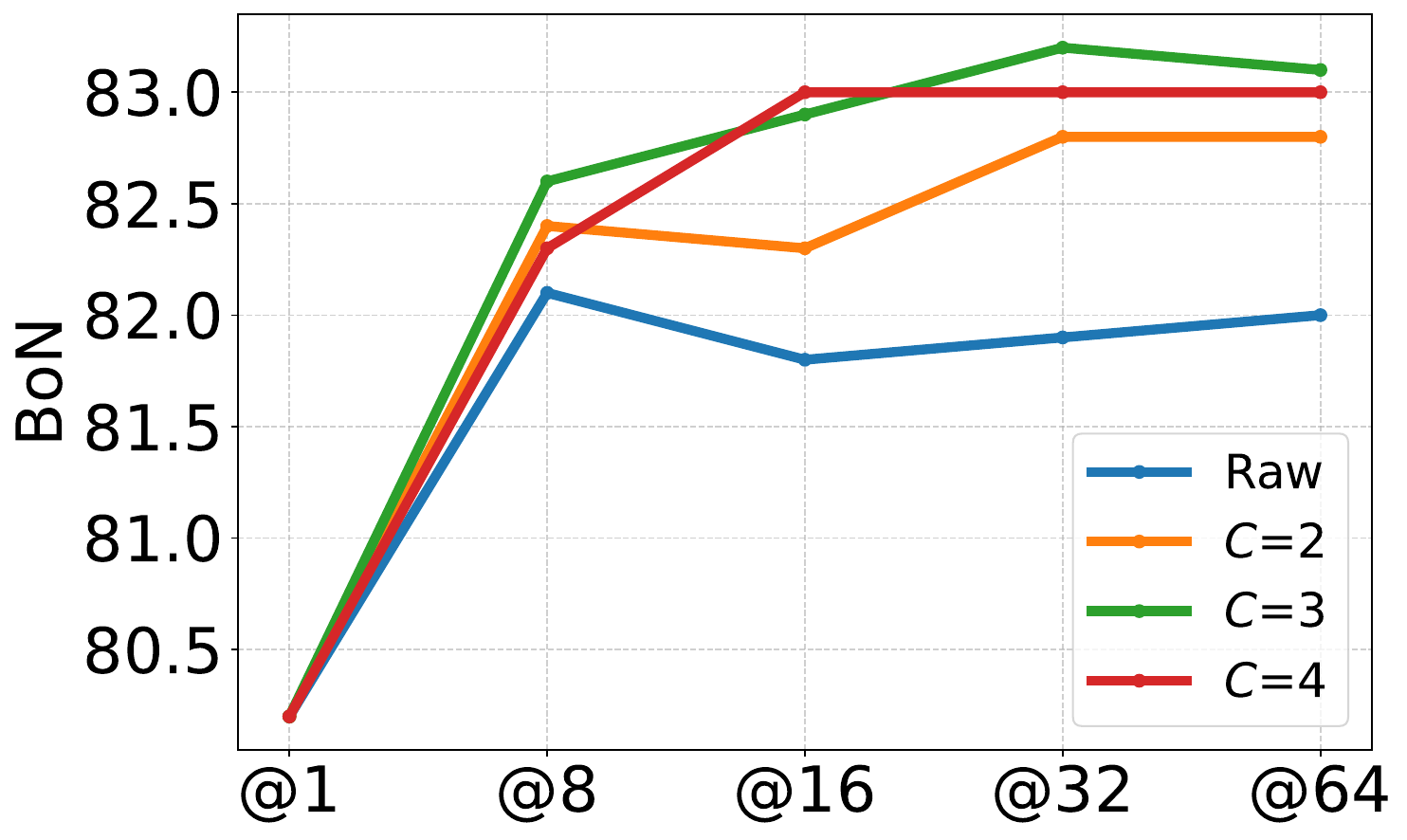}}      
  \subfigure[MSE on GSM-Plus.]{
    \label{subgraph_similarity}
    \includegraphics[width=0.23\textwidth]{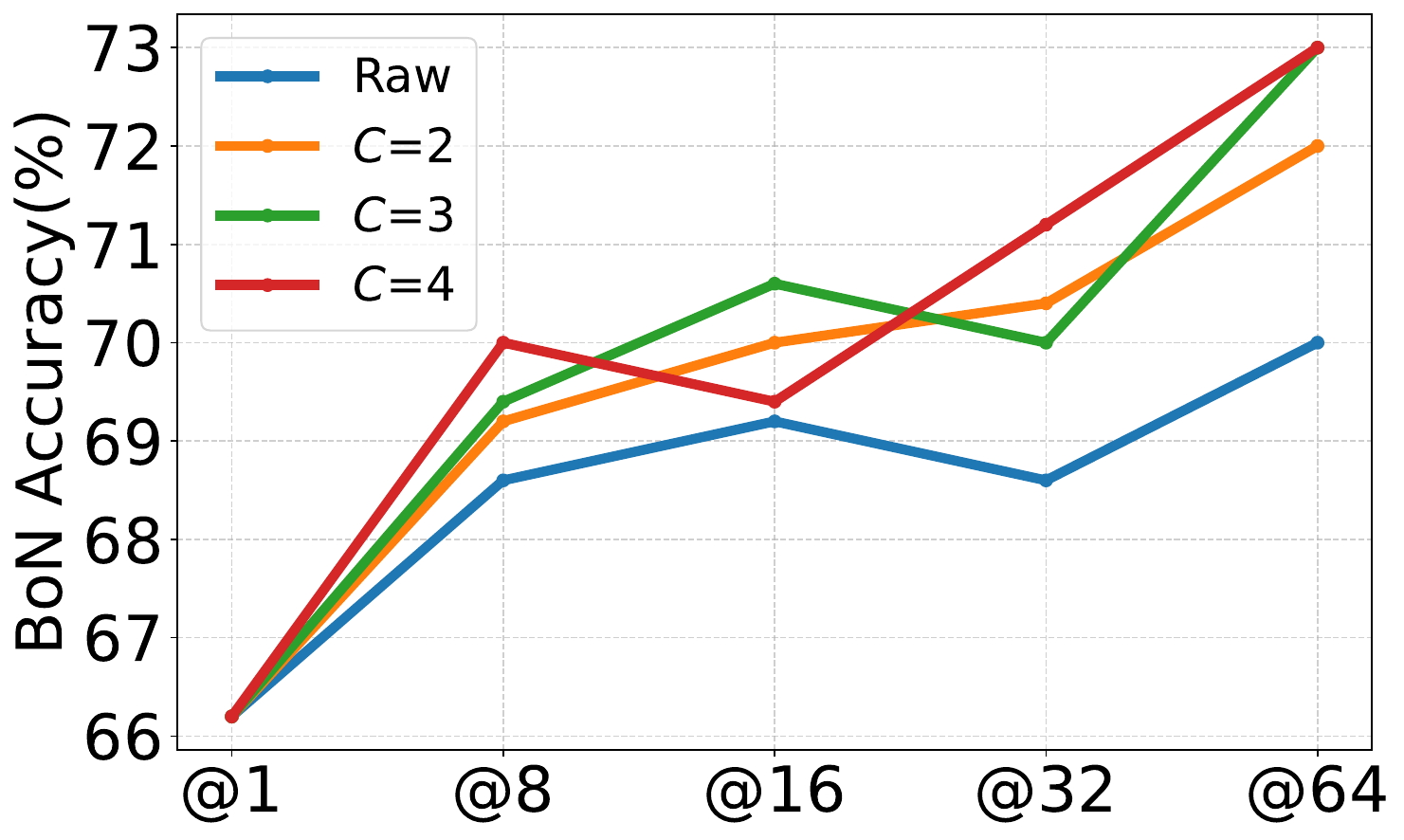}}  
  \subfigure[MSE on MATH500.]{
    \label{subgraph_similarity}
    \includegraphics[width=0.23\textwidth]{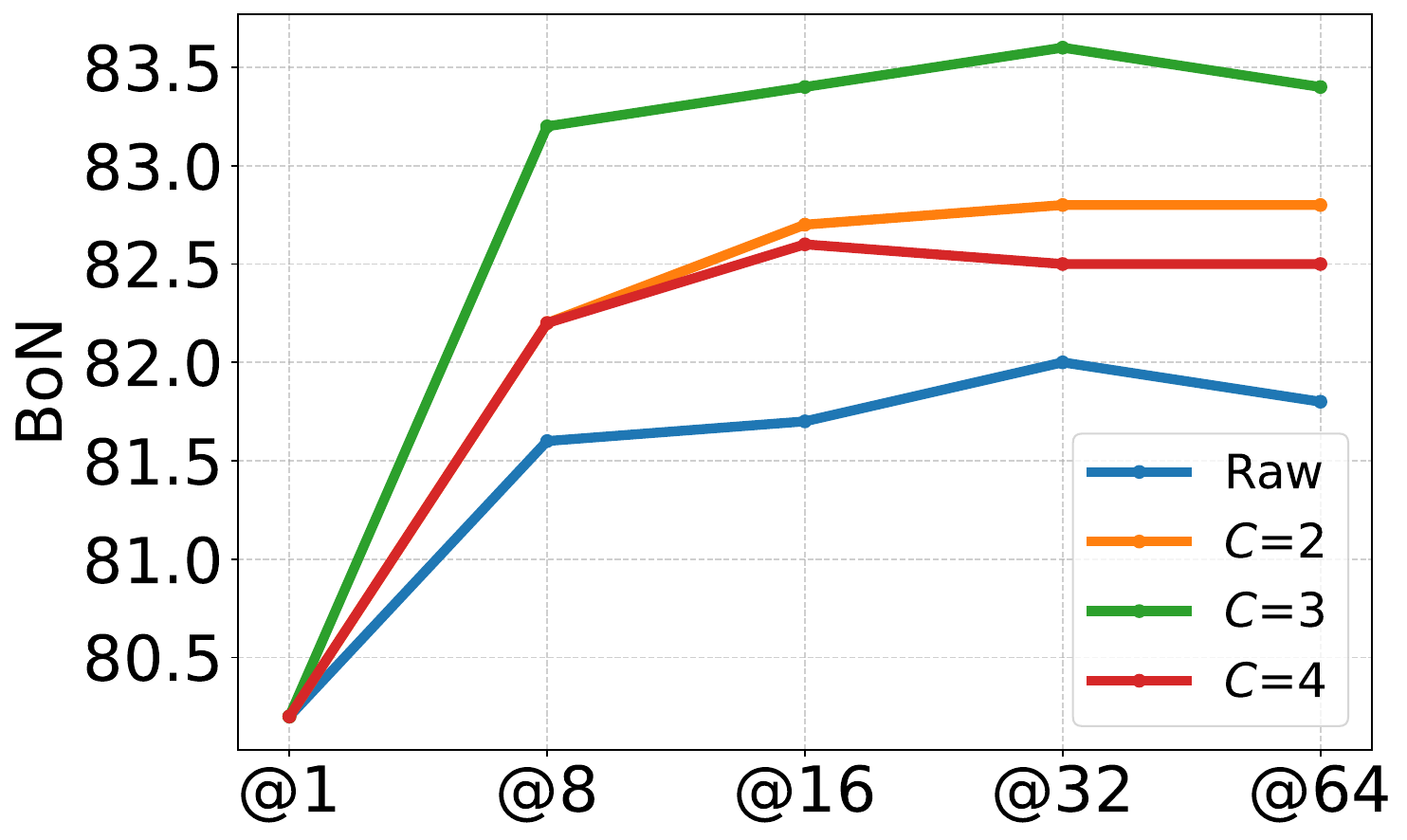}}      
  \subfigure[Q-Ranking on GSM-Plus.]{
    \label{fig:qranking_gsm}
    \includegraphics[width=0.23\textwidth]{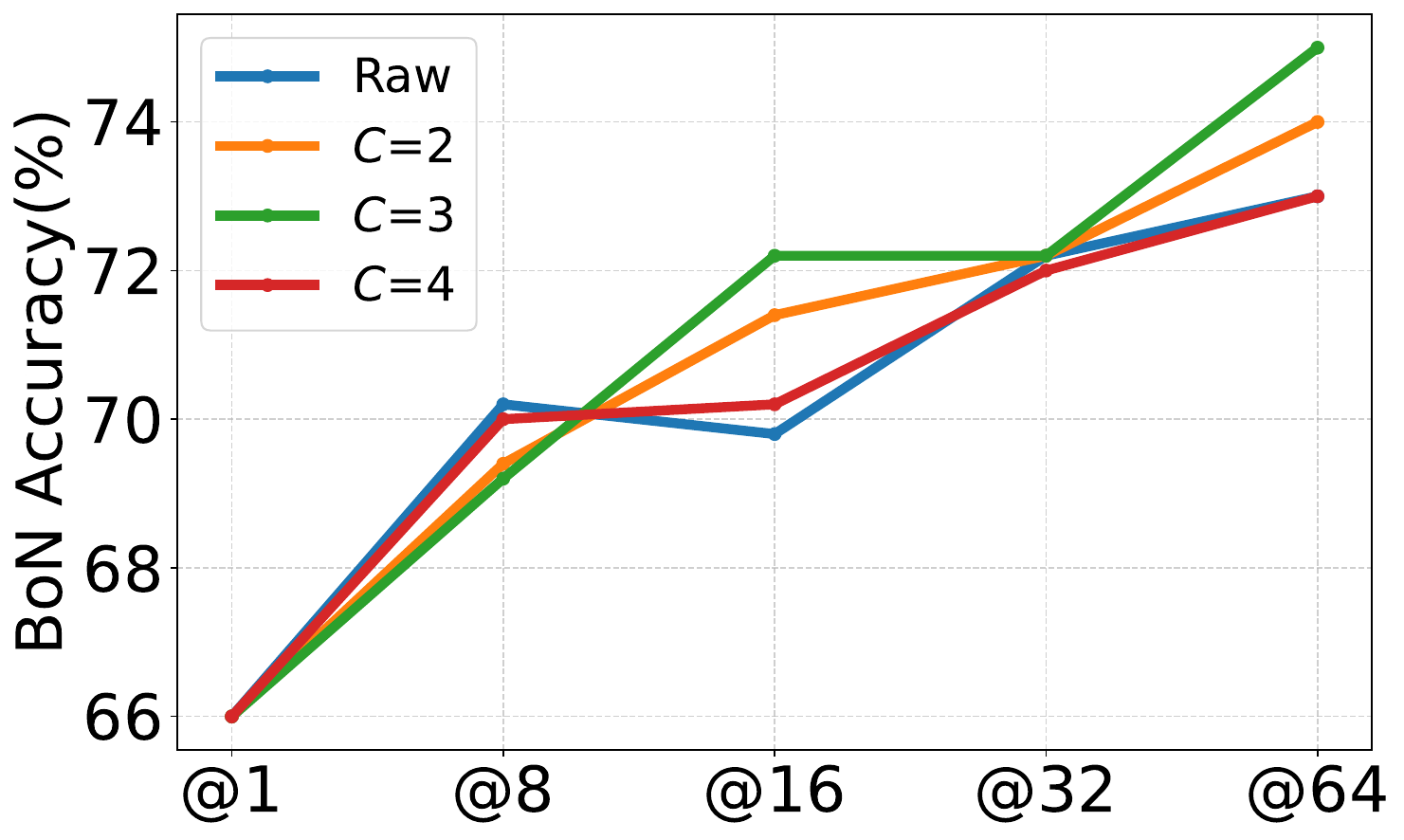}}  
  \subfigure[Q-Ranking on MATH500.]{
    \label{fig:qranking_math500}
    \includegraphics[width=0.23\textwidth]{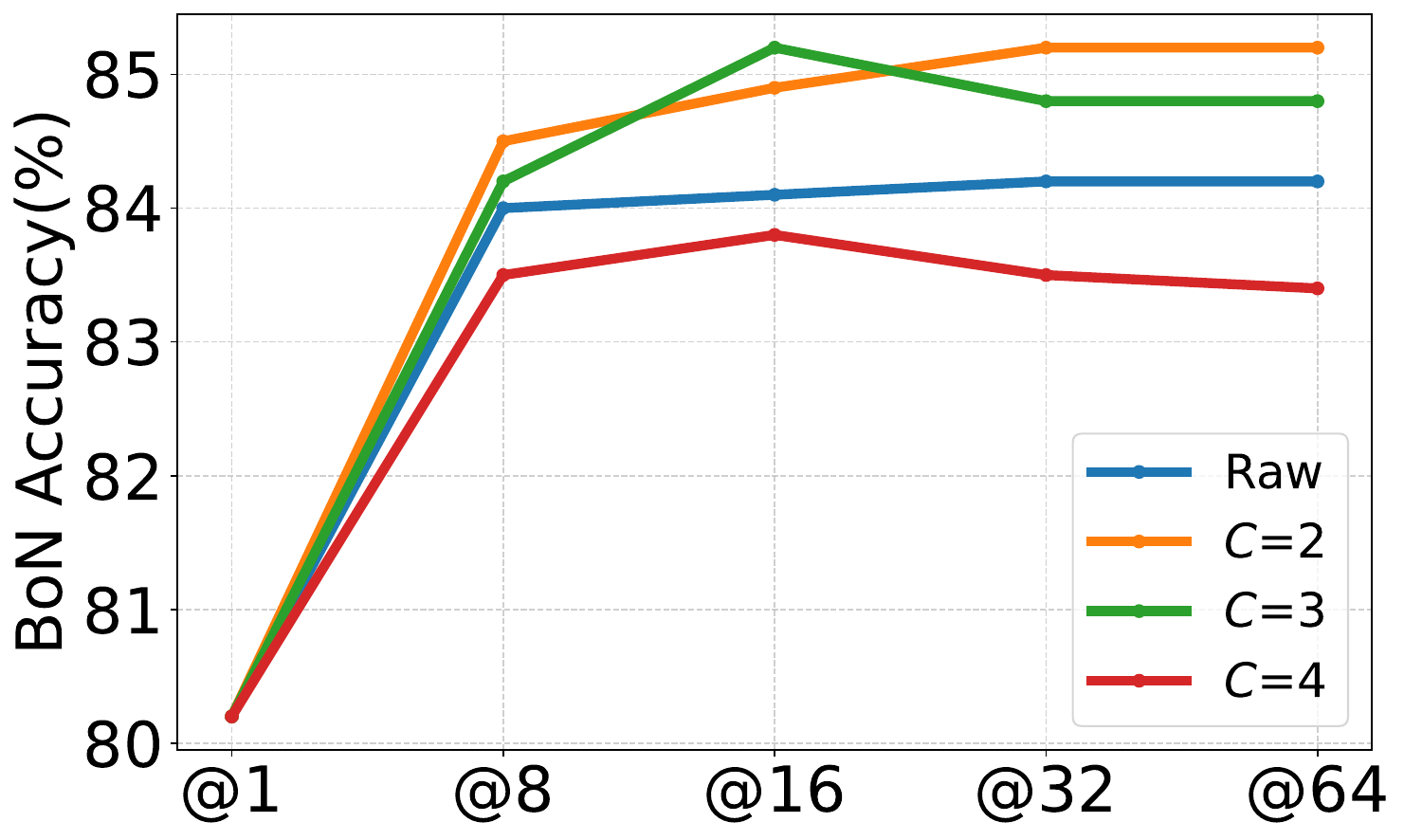}}      
  \caption{The BoN accuracy change under different values of $C$.} 
  \setlength{\abovecaptionskip}{0cm}
  \label{fig:performance_variation}
\end{figure}

\subsection{Main Results}
For \model, we set \( C \) to 2 to merge the adjacent two steps, showing the performance in Table~\ref{tab:main_results}. SHerd and RMCTS* are abbreviations for ShepHerd~\cite{shepherd} and ReSTMCTS*~\cite{restmcts}, respectively. The last column of each dataset indicates the average performance across four sampling conditions.

Our experimental results indicate that \model consistently brings performance improvements across all configurations, irrespective of the backbone model or loss objectives, underscoring its generalizability. For example, when employing MSE as the loss function, \model improves upon the baseline, ReSTMCTS*~\cite{restmcts}, by 1.3\% and 3.4\% on GSM-Plus and MATH500, respectively. Among the three learning objectives, the Q-ranking criterion exhibits superior performance, likely due to its foundation in the Markov Decision Process (MDP), which emphasizes evaluating transitions between adjacent steps. However, the presence of redundant steps can impede this learning process. By integrating \model with the Q-ranking-based method~\cite{pqm}, we observe further performance enhancements. Overall, these findings robustly affirm the efficacy and adaptability of \model. Considering the simplicity of \model, \model can be applied as a plug-and-play strategy to various scenarios.

\subsection{Further Studies}
We further explore the impact of varying \( C \), ranging from 2 to 4. It is worth noting that only the newly synthesized data is added. For simplicity, we only take the Qwen2.5-7B-MATH as the backbone model, leveraging BCE, MSE, and Q-ranking as the loss objectives, studying the performance changes on GSM-Plus and MATH500.

We can observe from Figure~\ref{fig:performance_variation} that \model generally brings performance gains across different learning objectives. However, the impact of different \( C \) varies, indicating that the optimal merge window size is not fixed for different learning criteria. In addition, we also find that the Q-ranking-based method is more sensitive to \( C \). When \( C \) is set to 4, the performance of the Q-ranking-based method on GSM-Plus does not exhibit a difference compared to the raw baseline (Figure~\ref{fig:qranking_gsm}). Moreover, the performance on MATH500 lags behind the raw baseline (Figure~\ref{fig:qranking_math500}). We ascribe this to the fact that the Q-ranking-based objective is sensitive to the interdependence between steps, and a large merging window hinders the learning of necessary fine-grained dependencies. Generally, setting \( C \) to 2 or 3 can better boost the overall ranking performance.

\section{Conclusion}
In this paper, we briefly review the issue of redundant steps in process data collection for PRM training, which may hinder downstream performance. To tackle this problem, we propose \model, a coarse-to-fine strategy that employs a sliding window to collect process data at diverse granularities. We validate \model across multiple experimental settings, confirming its effectiveness and versatility.
 
\newpage
\section{Limitations}
A more reasonable method should involve detecting and removing redundant steps, which we have not discussed. Another limitation is that the optimal $C$ should be designed adaptively with respect to different loss criteria. Future work should focus on designing methods to accurately detect redundant steps and developing adaptive methods to choose a feasible value of $C$.
% \newpage
\bibliography{ref}
\bibliographystyle{acl_natbib}

\appendix

\section{Supplement Material}
\begin{algorithm}[h]
\caption{Coarse-to-Fine Step Merging and Relabeling}
\label{algor}
\label{alg:coarse_to_fine}
\begin{algorithmic}[1]
\REQUIRE Trajectory of reasoning steps $S = \{s_1, s_2, \dots, s_N\}$, where each step $s_i$ has a label $l_i \in \{+, -\}$.
\ENSURE Mixed training corpus $\mathcal{D}$ with samples of diverse granularity.

\STATE Initialize $C_{\text{max}}$ as the maximum window size.
\STATE Initialize $\mathcal{D} \gets \emptyset$ \COMMENT{Empty set to store merged samples.}

\FOR{$C = C_{\text{max}}$ \TO $1$}
    \FOR{each window of size $C$ in $S$}
        \STATE Merge steps in the window into a single step $s_{i:j}$, where $j = i + C - 1$.
        \STATE Assign the label of $s_{i:j}$ as the label of the last step $s_j$.
        \STATE Add $(s_{i:j}, l_j)$ to $\mathcal{D}_C$.
    \ENDFOR
    \STATE Combine $\mathcal{D}_C$ with $\mathcal{D}$.
\ENDFOR
\RETURN $\mathcal{D}$ \COMMENT{Mixed training corpus with diverse granularity.}
\end{algorithmic}
\end{algorithm}

\end{document}